\documentclass{article}

\PassOptionsToPackage{numbers, compress}{natbib}
\usepackage[final]{nips_2016}

\usepackage[utf8]{inputenc} 
\usepackage[T1]{fontenc}    
\usepackage{hyperref}       
\usepackage{url}            
\usepackage{booktabs}       
\usepackage{amsfonts}       
\usepackage{nicefrac}       
\usepackage{microtype}      

\usepackage{pgfplots}
\usepackage{graphicx}
\usepackage{color}
\usepackage{tabularx}
\usepackage{amsmath}

\definecolor{RED}{rgb}{1,0,0}

\newcommand{\expect}[1]{{\mathbb{E}}{{\left[{{#1}}\right]}}}
\newcommand{\norm}[1]{{\mid\!\mid\!{#1}\!\mid\!\mid}}
\newcommand{\relu}[1]{{\text{ReLU}}{{\left[{{#1}}\right]}}}
\newcommand{\var}[1]{{\text{Var}}{{\left[{{#1}}\right]}}}

\title{Swapout: Learning an ensemble of deep architectures}

\author{
  Saurabh Singh, Derek Hoiem, David Forsyth \\
  Department of Computer Science\\
  University of Illinois, Urbana-Champaign\\
  \texttt{\{ss1, dhoiem, daf\}@illinois.edu} \\
}

\begin{document}

\maketitle


\begin{abstract}
We describe Swapout, a new stochastic training method, that outperforms ResNets
of identical network structure yielding impressive results on CIFAR-10 and
CIFAR-100. Swapout samples from a rich set of architectures including
dropout~\cite{dropout}, stochastic depth~\cite{stochdepth} and residual
architectures~\cite{resnets,resnetsv2} as special cases. When viewed as a
regularization method swapout not only inhibits co-adaptation of units in a
layer, similar to dropout, but also across network layers. We conjecture that
swapout achieves strong regularization by implicitly tying the parameters
across layers. When viewed as an ensemble training method, it samples a much
richer set of architectures than existing methods such as dropout or stochastic
depth. We propose a parameterization that reveals connections to exiting
architectures and suggests a much richer set of architectures to be explored.
We show that our formulation suggests an efficient training method and validate
our conclusions on CIFAR-10 and CIFAR-100 matching state of the art accuracy.
Remarkably, our 32 layer wider model performs similar to a 1001 layer ResNet
model.
\end{abstract}

\section{Introduction}

This paper describes swapout, a stochastic training method for general deep
networks. Swapout is a generalization of dropout~\cite{dropout} and stochastic
depth~\cite{stochdepth} methods. Dropout zeros the output of individual units
at random during training, while stochastic depth skips entire layers at random
during training. In comparison, the most general swapout network produces the
value of each output unit independently by reporting the sum of a randomly
selected subset of current and all previous layer outputs for that unit. As a
result, while some units in a layer may act like normal feedforward units,
others may produce skip connections and yet others may produce a sum of several
earlier outputs. In effect, our method averages over a very large set of
architectures that includes all architectures used by dropout and all used by
stochastic depth.

Our experimental work focuses on a version of swapout which is a natural
generalization of the residual network~\cite{resnets,resnetsv2}.  We show that
this results in improvements in accuracy over residual networks with the same
number of layers.

Improvements in accuracy are often sought by increasing the depth, leading to
serious practical difficulties.  The number of parameters rises sharply,
although recent works such as~\cite{vggnet,googlenet} have addressed this by
reducing the filter size~\cite{vggnet,googlenet}. Another issue resulting from
increased depth is the difficulty of training longer chains of dependent
variables. Such difficulties have been addressed by architectural innovations
that introduce shorter paths from input to loss either
directly~\cite{googlenet,highwaynet,resnets} or with additional losses applied
to intermediate layers~\cite{googlenet,lee2014deeply}. At the time of writing,
the deepest networks that have been successfully trained are residual networks
(1001 layers~\cite{resnetsv2}). We show that increasing the depth of our
swapout networks increases their accuracy.

There is compelling experimental evidence that these very large depths are
helpful,  though this may be because architectural innovations introduced to
make networks trainable reduce the capacity of the layers. The theoretical
evidence that a depth of 1000 is {\em  required} for practical problems is
thin. Bengio and Dellaleau argue that circuit efficiency constraints suggest
increasing depth is important, because there are functions that require
exponentially large shallow networks to compute~\cite{bandd}.   Less
experimental interest has been displayed in the width of the networks (the
number of filters in a convolutional layer). We show that increasing the width
of our swapout networks leads to significant improvements in their accuracy; an
appropriately wide swapout network is competitive with a deep residual network
that is 1.5 orders of magnitude deeper and has more parameters.

{\bf Contributions:} Swapout is a novel stochastic training scheme that can
sample from a rich set of architectures including dropout, stochastic depth and
residual architectures as special cases. Swapout improves the performance of
the residual networks for a model of the same depth. Wider but much shallower
swapout networks are competitive with very deep residual networks.

\section{Related Work}

Convolutional neural networks have a long history (see the introduction
of~\cite{lecunprocieee}). They are now intensively studied as a result of
recent successes (e.g.~\cite{alexnet}).   Increasing the number of layers in a
network improves performance~\cite{vggnet,googlenet} if the network can be
trained.  A variety of significant architectural innovations improve
trainability, including: the ReLU~\cite{nair2010rectified,glorot2011deep};
batch normalization~\cite{batchnorm}; and allowing signals to skip layers.

Our method exploits this skipping process.  Highway networks use gated skip
connections to allow information and gradients to pass unimpeded across several
layers~\cite{highwaynet}.  Residual networks use identity skip connections to
further improve training~\cite{resnets}; extremely deep residual networks can
be trained, and perform well~\cite{resnetsv2}.  In contrast to these
architectures, our method skips at the unit level (below), and does so
randomly.

Our method employs randomness at training time. For a review of the history of
random methods, see the introduction of~\cite{rahimirecht}, which shows that
entirely randomly chosen features can produce an SVM that generalizes well.
Randomly dropping out unit values (dropout~\cite{dropout}) discourages co-
adaptation between units. Randomly skipping layers (stochastic
depth)~\cite{stochdepth} during training reliably leads to improvements at test
time, likely  because doing so regularizes the network. The precise details of
the regularization remain uncertain, but it appears that stochastic depth
represents a form of tying between layers; when a layer is dropped, other
layers are encouraged to be able to replace it.  Each method can be seen as
training a network that averages over a family of architectures during
inference. Dropout averages over architectures with ``missing'' units and
stochastic depth averages over architectures with ``missing'' layers. Other
successful recent randomized methods include dropconnect~\cite{dropconnect}
which generalizes dropout by dropping individual connections instead of units
(so dropping several connections together), and stochastic
pooling~\cite{stochpooling} (which regularizes by replacing the deterministic
pooling by randomized pooling). In contrast, our method skips layers randomly
at a unit level  enjoying the benefits of each method.

Recent results show that (a) stochastic gradient descent with sufficiently few
steps is stable (in the sense that changes to training data do not unreasonably
disrupt predictions) and (b) dropout enhances that property, by reducing the
value of a Lipschitz constant (\cite{HardtRechtSinger}, Lemma 4.4).   We show
our method enjoys the same behavior as dropout in this framework.

Like dropout, the network trained with swapout depends on random variables.  A
reasonable strategy at test time with such a network is to evaluate multiple
instances (with different samples used for the random variables) and average.
Reliable improvements in accuracy are achievable by training distinct models
(which have {\em distinct} sets of parameters), then averaging
predictions~\cite{googlenet},  thereby forming an explicit ensemble. In
contrast, each of the instances of our network in an average would draw from
the {\em same} set of parameters (we call this an implicit ensemble).
Srivastava et al. argue that, at test time, random values in a dropout network
should be replaced with expectations, rather than taking an average over
multiple instances~\cite{dropout} (though they use explicit ensembles,
increasing the computational cost). Considerations include runtime at test; the
number of samples required; variance; and experimental accuracy results.    For
our model, accurate values of these expectations are not available.  In
Section~\ref{sec:experiments}, we show that (a) swapout networks that use
estimates of these expectations outperform strong comparable baselines and (b)
in turn, these are outperformed by swapout networks that use an implicit
ensemble.


\begin{figure*}[t]
  \centering
  \includegraphics[width=1.0\linewidth]{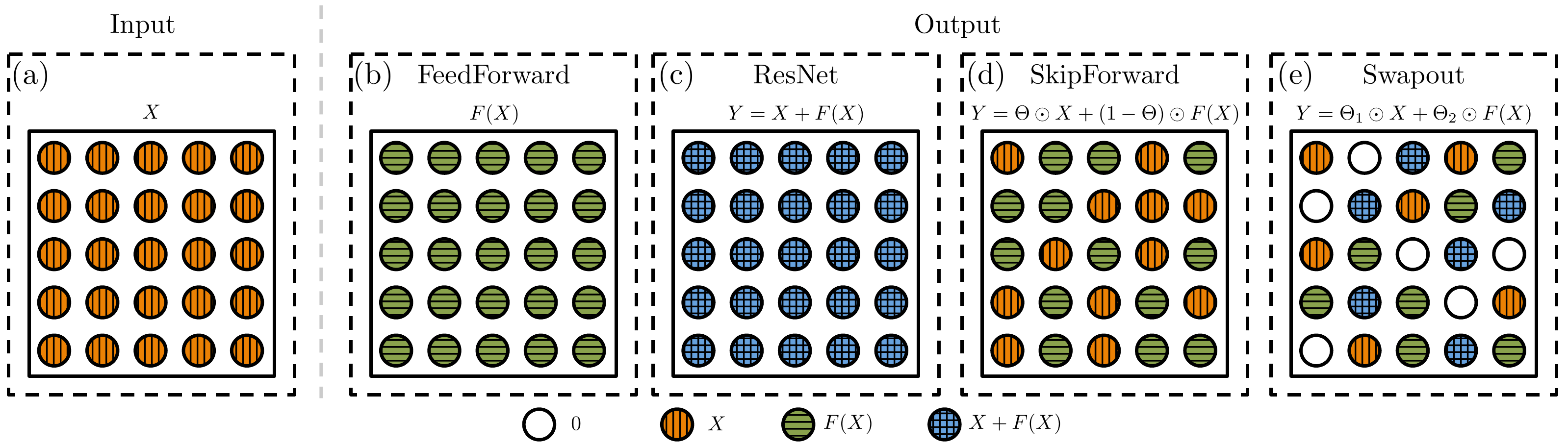}
  \caption{Visualization of architectural differences, showing
    computations for a block using various architectures.  Each circle is a
    unit in a grid corresponding to spatial layout, and circles are colored to
    indicate what they report. Given input $X$ ({\bf a}), all units in a feed
    forward block emit $F(X)$ ({\bf  b}). All units in a residual network block
    emit $X+F(X)$ ({\bf c}).  A skipforward network randomly chooses between
    reporting $X$ and $F(X)$ {\em per unit} ({\bf d}).  Finally, swapout
    randomly chooses between reporting $0$ (and so dropping out the unit), $X$
    (skipping the unit), $F(X)$ (imitating a feedforward network
    at the unit) and $X+F(X)$ (imitating a residual network unit).}
  \label{fig:arch_diff}
\end{figure*}

\section{Swapout}

\paragraph{Notation and terminology:}
We use capital letters to represent tensors and $\odot$ to represent element-
wise product (broadcasted for scalars). We use boldface $\mathbf{0}$ and
$\mathbf{1}$ to represent tensors of 0 and 1 respectively. A network block is a
set of simple layers in some specific configuration e.g. a convolution followed
by a ReLU or a residual network block~\cite{resnets}. Several such potentially
different blocks can be connected in the form of a directed acyclic graph to
form the full network model.

Dropout kills individual units randomly; stochastic depth skips entire blocks
of units randomly.  Swapout allows individual units to be dropped, or to skip
blocks randomly. Implementing swapout is a straightforward generalization of
dropout. Let $X$ be the input to some network block that computes $F(X)$.   The
$u$'th unit produces $F^{(u)}(X)$ as output. Let $\Theta$ be a tensor of i.i.d.
Bernoulli random variables. Dropout computes the output $Y$ of that block as
\begin{align}
\label{eq:dropout}
Y = \Theta \odot F(X).
\end{align}
It is natural to think of dropout as randomly selecting an output from the set
$\mathcal{F}^{(u)}=\{0, F^{(u)}(X)\}$ for the $u$'th unit.

Swapout generalizes dropout by expanding the choice of $\mathcal{F}^{(u)}$. Now
write $\{\Theta_i\}$ for $N$ distinct tensors of iid Bernoulli random variables
indexed by $i$ and with corresponding parameters $\{\theta_i\}$. Let $\{F_i\}$
be corresponding tensors consisting of values already computed somewhere in the
network. Note that one of these $F_i$ can be $X$ itself (identity). However,
$F_i$ are not restricted to being a function of $X$ and we drop the $X$ to
indicate this. Most natural choices for $F_i$ are the outputs of earlier
layers. Swapout computes the output of the layer in question by computing
\begin{align}
\label{eq:swapoutgeneral}
Y = \sum_{i=1}^N \Theta_i \odot F_i
\end{align}
and so, for unit $u$, we have $\mathcal{F}^{(u)}=\{F^{(u)}_1, F^{(u)}_2, \ldots, F^{(u)}_1+F^{(u)}_2, \ldots, \sum_i F^{(u)}_i\}$. We study the simplest case where
\begin{align}
\label{eq:swapout}
Y = \Theta_1 \odot X + \Theta_2 \odot F(X)
\end{align}
so that, for unit $u$, we have $\mathcal{F}^{(u)}=\{0, X^{(u)}, F^{(u)}(X), X^{(u)}+F^{(u)}(X)\}$.  Thus, each unit in the layer could be: 
\begin{description}
\item[1)] dropped (choose $0$);
\item[2)] a feedforward unit (choose $F^{(u)}(X)$);
\item[3)] skipped (choose $X^{(u)}$);
\item[4)] or a residual network unit (choose $X^{(u)}+F^{(u)}(X)$).
\end{description}
Since a swapout network can clearly imitate a residual network, and since
residual networks are currently the best-performing networks on various
standard benchmarks, we perform exhaustive experimental comparisons
with them.

If one accepts the view of dropout and stochastic depth as averaging over a set
of architectures, then swapout extends the set of architectures used.
Appropriate random choices of $\Theta_1$ and $\Theta_2$ yield: all
architectures covered by dropout; all architectures covered by stochastic
depth; and block level skip connections. But other choices yield unit level
skip and residual connections.

Swapout retains important properties of dropout.  Swapout
discourages co-adaptation by dropping units, but also by on occasion
presenting units with
inputs that have come from earlier layers. Dropout has been shown to enhance
the stability of stochastic gradient descent (\cite{HardtRechtSinger}, lemma
4.4). This applies to swapout in its most general form, too. We extend the
notation of that paper, and write $L$ for a Lipschitz constant that applies to
the network, $\nabla f(v)$ for the gradient of the network $f$ with parameters
$v$,  and $D\nabla f(v)$ for the gradient of the dropped out version of the
network.

The crucial point in the relevant enabling lemma is that $\expect{\norm{Df(v)}}
< \expect{\norm{\nabla f(v)}}\leq L$ (the inequality implies improvements).
Now write $\nabla S\left[f\right](v)$ for the gradient of a swapout network,
and $\nabla G\left[f\right](v)$ for the gradient of the swapout network which
achieves the largest Lipschitz constant by choice of $\Theta_i$ (this exists,
because $\Theta_i$ is discrete).  First, a Lipschitz constant applies to this
network; second, $\expect{\norm{\nabla S\left[f\right](v)}} \leq
\expect{\norm{\nabla G\left[f\right](v)}}\leq L$, so swapout makes stability no
worse; third, we speculate light conditions on $f$ should provide
$\expect{\norm{\nabla S\left[f\right](v)}} <\expect{\norm{\nabla
G\left[f\right](v)}}\leq L$, improving stability
(\cite{HardtRechtSinger}~Section~4).




\subsection{Inference in Stochastic Networks} 

A model trained with swapout represents an entire family of networks with tied
parameters, where members of the family were sampled randomly during training.
There are two options for inference.  We could either replace random variables
with their expected values, as recommended by Srivastava et al.~\cite{dropout}
(deterministic inference).  Alternatively, we could sample several members of
the family at random, and average their predictions (stochastic inference).

There is an important difference between swapout and dropout.  In a dropout
network, one can estimate expectations exactly (as long as the network isn't
trained with batch normalization, below).  This is because
$\expect{\relu{\Theta
    \odot F(X)}}=\relu{\expect{\Theta \odot F(X)}}$ (recall $\Theta$ is a
tensor of Bernoulli random variables, and thus non-negative). 

In a swapout network, one usually can not estimate expectations exactly. The
problem is that $\expect{\relu{(\Theta_1 X + \Theta_2 Y)}}$ is not the same as
$\relu{\expect{(\Theta_1 X + \Theta_2 Y)}}$ in general.  Estimates of
expectations that ignore this are successful, as the experiments show,
but stochastic inference gives significantly better results.

Srivastava et al. argue that deterministic inference is significantly less
expensive in computation. We believe that Srivastava et al. may have
overestimated how many samples are required for an accurate average, because
they use {\em distinct} dropout networks in the average (Figure 11
in~\cite{dropout}).  Our experience of stochastic inference with swapout has
been positive, with the number of samples needed for good behavior small
(Figure~\ref{fig:sample_estimates}). Furthermore, computational costs of
inference are smaller when each instance of the network uses the same
parameters


A technically more delicate point is that both dropout and swapout networks interact poorly with batch normalization
{\em if} one uses deterministic inference. 
The problem is that the estimates collected by batch normalization during
training may not reflect test time statistics. To see this consider two random
variables $X$ and $Y$ and let
$\Theta_1, \Theta_2 \sim Bernoulli(\theta)$. While $\expect{\Theta_1 X +
\Theta_2 Y} =\expect{\theta X + \theta Y} = \theta X + \theta Y$, it can be
shown that $\var{\Theta_1 X + \Theta_2 Y} \ge \var{\theta X + \theta Y}$ with
equality holding only for $\theta=0$ and $\theta=1$. Thus, the variance
estimates collected by Batch Normalization during training do not represent the
statistics observed during testing if the expected values of $\Theta_1$
and $\Theta_2$ are used in a deterministic inference scheme. These errors in
scale estimation accumulate as more and more layers are stacked. This may
explain why~\cite{stochdepth} reports that dropout doesn't lead to any
improvement when used in residual networks with batch normalization.



\subsection{Baseline comparison methods}

\paragraph{ResNets:}
We compare with ResNet architectures as described in~\cite{resnets}(referred to
as v1) and in~\cite{resnetsv2}(referred to as v2).

\paragraph{Dropout:}
We use standard dropout (replace equation~\ref{eq:swapout} with equation~\ref{eq:dropout}).

\paragraph{Layer Dropout:}
We replace equation~\ref{eq:swapout} by $Y = X + \Theta^{(1 \times 1)} 
F(X)$. Here $\Theta^{(1 \times 1)}$ is a single Bernoulli
random variable shared across all units.

\paragraph{SkipForward:}
Equation~\ref{eq:swapout} introduces two stochastic parameters $\Theta_1$ and
$\Theta_2$. We also explore and compare with a simpler architecture,
SkipForward, that introduces only one parameter but samples from a smaller set
$\mathcal{F}^{(u)} = \{X^{(u)}, F^{(u)}(X)\}$ as below.

\begin{align}
\label{eq:skipforward}
Y = \Theta \odot X + (1 - \Theta) \odot F(X)
\end{align}

\section{Experiments}
\label{sec:experiments}

We experiment extensively on the CIFAR-10 dataset and demonstrate that a model
trained with swapout outperforms a comparable ResNet model. Further, a 32 layer
wider model matches the performance of a 1001 layer ResNet on both CIFAR-10 and
CIFAR-100 datasets.

\paragraph{Model:}
We experiment with ResNet architectures as described in~\cite{resnets}(referred
to as v1) and in~\cite{resnetsv2}(referred to as v2). However, our
implementation (referred to as ResNet Ours) has the following modifications
which improve the performance of the original model
(Table~\ref{tab:cifar1020layerbase}). Between blocks of different feature sizes
we subsample using average pooling instead of strided convolutions and use
projection shortcuts with learned parameters. For final prediction we follow a
scheme similar to Network in Network~\cite{netinnet}. We replace average
pooling and fully connected layer by a 1x1 convolution layer followed
by global average pooling to predict the logits that are fed into the softmax.

Layers in ResNets are arranged in three groups with all convolutional layers in
a group containing equal number of filters. We represent the number of filters
in each group as a tuple with the smallest size as (16, 32, 64) (as used
in~\cite{resnets}for CIFAR-10). We refer to this as \emph{width} and experiment
with various multiples of this base size represented as $W \times 1$, $W \times
2$ etc.

\paragraph{Training:}
We train using SGD with a batch size of 128, momentum of 0.9 and weight decay
of 0.0001. Unless otherwise specified, we train all the models for a total 256
epochs. Starting from an initial learning rate of 0.1, we drop it by a factor
of 10 after 196 epochs and then again after 224 epochs. We do the standard
augmentation of left-right flips and random translations of up to four pixels.
For translation, we pad the images by 4 pixels on all the sides and sample
a random 32x32 crop. All the images in a mini-batch use the same crop. Note
that dropout slows convergence (\cite{dropout}, A.4), and swapout should do so
too for similar reasons. Thus using the same training schedule for all the
methods should \emph{disadvantage} swapout.

\begin{table}[t]
  \caption{In comparison with fair baselines on CIFAR-10, swapout is always
  more accurate. We refer to the base width of (16, 32, 64) as $W \times 1$ and
  others are multiples of it (See Table~\ref{tab:cifar10width} for details on
  width). We report the width along with the number of parameters in each
  model. Models trained with swapout consistently outperform all other models
  of comparable architecture. All stochastic methods were trained using the
  Linear(1, 0.5) schedule (Table~\ref{tab:stochastic_eval}). v1 and v2
  represent residual block architectures in~\cite{resnets} and~\cite{resnetsv2}
  respectively.}
  \label{tab:cifar1020layerbase}
  \centering
  \begin{tabular}{llll}
    \toprule
    Method   & Width & \#Params & Error(\%)   \\
    \midrule
    ResNet v1~\cite{resnets} & $W \times 1$ & 0.27M & 8.75 \\
    ResNet v1 Ours           & $W \times 1$ & 0.27M & 8.54 \\
    Swapout v1               & $W \times 1$ & 0.27M & \textbf{8.27} \\
    \midrule
    ResNet v2 Ours           & $W \times 1$ & 0.27M & 8.27 \\
    Swapout v2               & $W \times 1$ & 0.27M & \textbf{7.97} \\
    \midrule
    Swapout v1               & $W \times 2$ & 1.09M & 6.58 \\
    \midrule
    ResNet v2 Ours           & $W \times 2$ & 1.09M & 6.54 \\
    Stochastic Depth v2 Ours & $W \times 2$ & 1.09M & 5.99 \\
    Dropout v2               & $W \times 2$ & 1.09M & 5.87 \\
    SkipForward v2           & $W \times 2$ & 1.09M & 6.11 \\
    Swapout v2               & $W \times 2$ & 1.09M & \textbf{5.68} \\
    \bottomrule
  \end{tabular}
\end{table}

\begin{table}[t]
  \caption{The choice of stochastic training schedule matters. We evaluate the
  performance of a 20 layer swapout model ($W \times 2$) trained with different stochasticity
  schedules on CIFAR-10. These schedules differ in how the parameters $\theta_1$ and
  $\theta_2$ of the Bernoulli random variables in equation~\ref{eq:swapout} are
  set for the different layers. Linear($a$, $b$) refers to linear interpolation
  from $a$ to $b$ from the first block to the last (see~\cite{stochdepth}). Others use the same value
  for all the blocks. We report the performance for both the deterministic and
  stochastic inference (with 30 samples). Schedule with less randomness in the
  early layers (bottom row) performs the best.}
  \label{tab:stochastic_eval}
  \centering
  \begin{tabular}{lll}
    \toprule
    Method     & Deterministic Error(\%) & Stochastic Error(\%) \\
    \midrule
    Swapout ($\theta_1=\theta_2=0.5$)                 & 10.36 & 6.69 \\
    Swapout ($\theta_1=0.2, \theta_2=0.8$)            & 10.14 & 7.63 \\
    Swapout ($\theta_1=0.8, \theta_2=0.2$)            & 7.58  & 6.56 \\
    Swapout ($\theta_1=\theta_2=$ Linear(0.5, 1))     & 7.34  & 6.52 \\
    Swapout ($\theta_1=\theta_2=$ Linear(1, 0.5))     & \textbf{6.43}   & \textbf{5.68} \\
    \bottomrule
  \end{tabular}
\end{table}

\paragraph{Models trained with Swapout consistently outperform baselines:}
Table~\ref{tab:cifar1020layerbase} compares Swapout with various 20 layer
baselines. Models trained with Swapout consistently outperform all other
models of similar architecture.

\paragraph{The stochastic training schedule matters:}
Different layers in a swapout network could be trained with different
parameters of their Bernoulli distributions (the stochastic training schedule).
Table~\ref{tab:stochastic_eval} shows that different stochastic training
schedules have a significant affect on the performance. We report the
performance with deterministic as well as stochastic inference. These schedules
differ in how the values of parameters $\theta_1$ and $\theta_2$ of the
Bernoulli random variables in equation~\ref{eq:swapout} are set for the
different layers. Note that $\theta_1=\theta_2=0.5$ corresponds to the maximum
stochasticity. A schedule with less randomness in the early layers (bottom row)
performs the best. This is expected because Swapout adds per unit noise and
early layers have the largest number of units. Thus, low stochasticity in early
layers significantly reduces the randomness in the system. We use this schedule for
all the experiments unless otherwise stated.

\paragraph{Swapout improves over ResNet architecture:}
From Table~\ref{tab:cifar10width} it is evident that networks trained with
Swapout consistently show better performance than corresponding ResNets, for
most choices of width investigated, using just the deterministic inference.
This difference indicates that the performance improvement is not just an
ensemble effect.

\paragraph{Stochastic inference outperforms deterministic inference:}
Table~\ref{tab:cifar10width} shows that the stochastic inference scheme
outperforms the deterministic scheme in all the experiments. Prediction for
each image is done by averaging the results of 30 stochastic forward passes.
This difference is not just due to the widely reported effect that an ensemble
of networks is better as networks in our ensemble share parameters. Instead,
stochastic inference produces more accurate expectations and interacts better
with batch normalization.

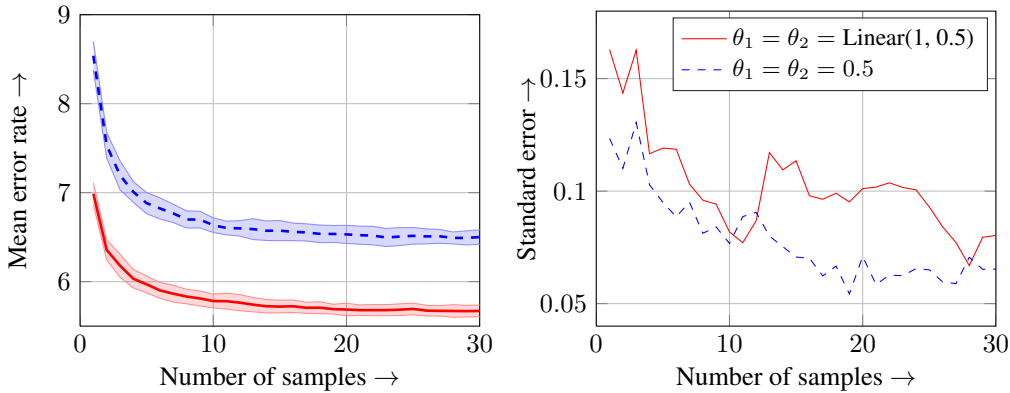
\begin{figure}[t]
\begin{center}
%
\begin{tikzpicture}

\begin{axis}[%
width=0.380368\linewidth,
height=0.296615\linewidth,
at={(0\linewidth,0\linewidth)},
scale only axis,
xmin=0,
xmax=30,
xlabel={Number of samples $\rightarrow$},
xmajorgrids,
ymin=5.5,
ymax=9,
ylabel={Mean error rate $\rightarrow$},
ymajorgrids,
xlabel style={yshift=0.1cm},
ylabel style={yshift=-0.4cm},
xticklabel style={/pgf/number format/.cd,precision=2,fixed},
yticklabel style={/pgf/number format/.cd,precision=2,fixed},
legend style={font=\footnotesize}
]

\addplot[area legend,solid,fill=red,opacity=1.500000e-01,draw=none,forget plot]
table[row sep=crcr] {%
x	y\\
1	6.86388473854366\\
2	6.2465657029006\\
3	6.04983471690814\\
4	5.9298745355368\\
5	5.87492108540797\\
6	5.81272432952693\\
7	5.76956204591202\\
8	5.75106491661714\\
9	5.72728043993345\\
10	5.70557903558903\\
11	5.69130576869565\\
12	5.67309502528691\\
13	5.66098541799563\\
14	5.64799705020526\\
15	5.64898034864432\\
16	5.65312728737652\\
17	5.64334046386597\\
18	5.64065333466641\\
19	5.63700511271499\\
20	5.61570501818274\\
21	5.62115835766514\\
22	5.61697784079723\\
23	5.61724427040093\\
24	5.61948918664325\\
25	5.62830342568394\\
26	5.61468529154746\\
27	5.61278354325921\\
28	5.5997413917095\\
29	5.60306040207119\\
30	5.60566241510434\\
30	5.73633758489569\\
29	5.73360626459553\\
28	5.74092527495716\\
27	5.7305497900742\\
26	5.7339813751192\\
25	5.75836324098273\\
24	5.7505108133568\\
23	5.74208906293239\\
22	5.74235549253611\\
21	5.73817497566826\\
20	5.75762831515058\\
19	5.74566155395169\\
18	5.77401333200028\\
17	5.7679928694674\\
16	5.79353937929013\\
15	5.790352984689\\
14	5.79866961646144\\
13	5.82101458200436\\
12	5.85423830804641\\
11	5.86869423130432\\
10	5.85908763107764\\
9	5.89538622673318\\
8	5.9136017500495\\
7	5.95977128742135\\
6	5.98994233713965\\
5	6.06507891459204\\
4	6.13545879779656\\
3	6.31083194975852\\
2	6.46676763043278\\
1	7.1107819281231\\
}--cycle;

\addplot [color=white!55!red,solid,forget plot]
  table[row sep=crcr]{%
1	6.86388473854366\\
2	6.2465657029006\\
3	6.04983471690814\\
4	5.9298745355368\\
5	5.87492108540797\\
6	5.81272432952693\\
7	5.76956204591202\\
8	5.75106491661714\\
9	5.72728043993345\\
10	5.70557903558903\\
11	5.69130576869565\\
12	5.67309502528691\\
13	5.66098541799563\\
14	5.64799705020526\\
15	5.64898034864432\\
16	5.65312728737652\\
17	5.64334046386597\\
18	5.64065333466641\\
19	5.63700511271499\\
20	5.61570501818274\\
21	5.62115835766514\\
22	5.61697784079723\\
23	5.61724427040093\\
24	5.61948918664325\\
25	5.62830342568394\\
26	5.61468529154746\\
27	5.61278354325921\\
28	5.5997413917095\\
29	5.60306040207119\\
30	5.60566241510434\\
};
\addplot [color=white!55!red,solid,forget plot]
  table[row sep=crcr]{%
1	7.1107819281231\\
2	6.46676763043278\\
3	6.31083194975852\\
4	6.13545879779656\\
5	6.06507891459204\\
6	5.98994233713965\\
7	5.95977128742135\\
8	5.9136017500495\\
9	5.89538622673318\\
10	5.85908763107764\\
11	5.86869423130432\\
12	5.85423830804641\\
13	5.82101458200436\\
14	5.79866961646144\\
15	5.790352984689\\
16	5.79353937929013\\
17	5.7679928694674\\
18	5.77401333200028\\
19	5.74566155395169\\
20	5.75762831515058\\
21	5.73817497566826\\
22	5.74235549253611\\
23	5.74208906293239\\
24	5.7505108133568\\
25	5.75836324098273\\
26	5.7339813751192\\
27	5.7305497900742\\
28	5.74092527495716\\
29	5.73360626459553\\
30	5.73633758489569\\
};
\addplot [color=red,solid,line width=1.0pt,forget plot]
  table[row sep=crcr]{%
1	6.98733333333338\\
2	6.35666666666669\\
3	6.18033333333333\\
4	6.03266666666668\\
5	5.97000000000001\\
6	5.90133333333329\\
7	5.86466666666669\\
8	5.83233333333332\\
9	5.81133333333332\\
10	5.78233333333334\\
11	5.77999999999998\\
12	5.76366666666666\\
13	5.74099999999999\\
14	5.72333333333335\\
15	5.71966666666666\\
16	5.72333333333332\\
17	5.70566666666668\\
18	5.70733333333335\\
19	5.69133333333334\\
20	5.68666666666666\\
21	5.6796666666667\\
22	5.67966666666667\\
23	5.67966666666666\\
24	5.68500000000003\\
25	5.69333333333334\\
26	5.67433333333333\\
27	5.6716666666667\\
28	5.67033333333333\\
29	5.66833333333336\\
30	5.67100000000002\\
};

\addplot[area legend,solid,fill=blue,opacity=1.500000e-01,draw=none,forget plot]
table[row sep=crcr] {%
x	y\\
1	8.37574064391741\\
2	7.38902513754335\\
3	7.03336339075514\\
4	6.89205525117821\\
5	6.7628810118691\\
6	6.70403745909582\\
7	6.66353783677145\\
8	6.60268981760631\\
9	6.60313732020702\\
10	6.55278328075176\\
11	6.52559815457804\\
12	6.51188213143396\\
13	6.47286759628521\\
14	6.46256508783758\\
15	6.45986282811037\\
16	6.46139003908157\\
17	6.45695445136799\\
18	6.43762683180772\\
19	6.44079495811672\\
20	6.43021514066314\\
21	6.42020281601797\\
22	6.41363719604322\\
23	6.39436011281653\\
24	6.4060905484499\\
25	6.42172462977151\\
26	6.42453449551779\\
27	6.43217836661774\\
28	6.42441381284624\\
29	6.41184608767579\\
30	6.42101175743026\\
30	6.58165490923643\\
29	6.57082057899087\\
28	6.55825285382042\\
27	6.58648830004898\\
26	6.59279883781552\\
25	6.60760870356183\\
24	6.60724278488345\\
23	6.59763988718346\\
22	6.62102947062343\\
21	6.62379718398198\\
20	6.63245152600356\\
19	6.63120504188331\\
18	6.63570650152563\\
17	6.64971221529867\\
16	6.65727662758511\\
15	6.68680383855631\\
14	6.68143491216245\\
13	6.70713240371474\\
12	6.68545120189936\\
11	6.67973517875529\\
10	6.71655005258156\\
9	6.79152934645961\\
8	6.79464351572703\\
7	6.86979549656188\\
6	6.94129587423753\\
5	7.00111898813094\\
4	7.12527808215513\\
3	7.35863660924486\\
2	7.67630819579002\\
1	8.70159268941591\\
}--cycle;

\addplot [color=white!55!blue,solid,forget plot]
  table[row sep=crcr]{%
1	8.37574064391741\\
2	7.38902513754335\\
3	7.03336339075514\\
4	6.89205525117821\\
5	6.7628810118691\\
6	6.70403745909582\\
7	6.66353783677145\\
8	6.60268981760631\\
9	6.60313732020702\\
10	6.55278328075176\\
11	6.52559815457804\\
12	6.51188213143396\\
13	6.47286759628521\\
14	6.46256508783758\\
15	6.45986282811037\\
16	6.46139003908157\\
17	6.45695445136799\\
18	6.43762683180772\\
19	6.44079495811672\\
20	6.43021514066314\\
21	6.42020281601797\\
22	6.41363719604322\\
23	6.39436011281653\\
24	6.4060905484499\\
25	6.42172462977151\\
26	6.42453449551779\\
27	6.43217836661774\\
28	6.42441381284624\\
29	6.41184608767579\\
30	6.42101175743026\\
};
\addplot [color=white!55!blue,solid,forget plot]
  table[row sep=crcr]{%
1	8.70159268941591\\
2	7.67630819579002\\
3	7.35863660924486\\
4	7.12527808215513\\
5	7.00111898813094\\
6	6.94129587423753\\
7	6.86979549656188\\
8	6.79464351572703\\
9	6.79152934645961\\
10	6.71655005258156\\
11	6.67973517875529\\
12	6.68545120189936\\
13	6.70713240371474\\
14	6.68143491216245\\
15	6.68680383855631\\
16	6.65727662758511\\
17	6.64971221529867\\
18	6.63570650152563\\
19	6.63120504188331\\
20	6.63245152600356\\
21	6.62379718398198\\
22	6.62102947062343\\
23	6.59763988718346\\
24	6.60724278488345\\
25	6.60760870356183\\
26	6.59279883781552\\
27	6.58648830004898\\
28	6.55825285382042\\
29	6.57082057899087\\
30	6.58165490923643\\
};
\addplot [color=blue,dashed,line width=1.0pt,forget plot]
  table[row sep=crcr]{%
1	8.53866666666666\\
2	7.53266666666669\\
3	7.196\\
4	7.00866666666667\\
5	6.88200000000002\\
6	6.82266666666668\\
7	6.76666666666667\\
8	6.69866666666667\\
9	6.69733333333332\\
10	6.63466666666666\\
11	6.60266666666666\\
12	6.59866666666666\\
13	6.58999999999997\\
14	6.57200000000002\\
15	6.57333333333334\\
16	6.55933333333334\\
17	6.55333333333333\\
18	6.53666666666668\\
19	6.53600000000002\\
20	6.53133333333335\\
21	6.52199999999998\\
22	6.51733333333333\\
23	6.496\\
24	6.50666666666667\\
25	6.51466666666667\\
26	6.50866666666666\\
27	6.50933333333336\\
28	6.49133333333333\\
29	6.49133333333333\\
30	6.50133333333335\\
};
\end{axis}
\end{tikzpicture}%
%
\begin{tikzpicture}

\begin{axis}[%
width=0.380368\linewidth,
height=0.3\linewidth,
at={(0\linewidth,0\linewidth)},
scale only axis,
xmin=0,
xmax=30,
xlabel={Number of samples $\rightarrow$},
xmajorgrids,
ymin=0.04,
ymax=0.18,
ylabel={Standard error $\rightarrow$},
ymajorgrids,
legend style={legend cell align=left,align=left,draw=white!15!black},
xlabel style={yshift=0.1cm},
ylabel style={yshift=-0.3cm},
xticklabel style={/pgf/number format/.cd,precision=2,fixed},
yticklabel style={/pgf/number format/.cd,precision=2,fixed},
ytick={0.05, 0.1, ..., 0.2},
legend style={font=\footnotesize}
]
\addplot [color=red,solid]
  table[row sep=crcr]{%
1	0.16292602274925\\
2	0.143641529123332\\
3	0.162636609244863\\
4	0.116611415488458\\
5	0.119118988130917\\
6	0.118629207570854\\
7	0.103128829895213\\
8	0.0959768490603631\\
9	0.0941960131262937\\
10	0.0818833859149009\\
11	0.0770685120886291\\
12	0.086784535232699\\
13	0.11713240371477\\
14	0.109434912162433\\
15	0.113470505222968\\
16	0.0979432942517709\\
17	0.0963788819653388\\
18	0.0990398348589555\\
19	0.0952050418832947\\
20	0.101118192670207\\
21	0.101797183982007\\
22	0.103696137290108\\
23	0.101639887183462\\
24	0.10057611821678\\
25	0.0929420368951632\\
26	0.0841321711488639\\
27	0.0771549667156207\\
28	0.0669195204870916\\
29	0.0794872456575415\\
30	0.0803215759030856\\
};
\addlegendentry{$\theta_1=\theta_2=\text{Linear(1, 0.5)}$};

\addplot [color=blue,dashed]
  table[row sep=crcr]{%
1	0.123448594789719\\
2	0.110100963766091\\
3	0.13049861642519\\
4	0.102792131129879\\
5	0.0950789145920367\\
6	0.0886090038063582\\
7	0.0951046207546646\\
8	0.081268416716183\\
9	0.0840528933998653\\
10	0.0767542977443092\\
11	0.0886942313043343\\
12	0.0905716413797515\\
13	0.0800145820043649\\
14	0.0753362831280892\\
15	0.0706863180223411\\
16	0.0702060459568039\\
17	0.0623262028007132\\
18	0.0666799986669338\\
19	0.0543282206183464\\
20	0.0709616484839226\\
21	0.0585083090015621\\
22	0.0626888258694407\\
23	0.0624223962657284\\
24	0.0655108133567791\\
25	0.0650299076493958\\
26	0.059648041785872\\
27	0.0588831234074943\\
28	0.0705919416238305\\
29	0.0652729312621673\\
30	0.0653375848956781\\
};
\addlegendentry{$\theta_1=\theta_2=0.5$};

\end{axis}
\end{tikzpicture}%
\end{center}
   \caption{Stochastic inference needs few samples for a good estimate. We plot
   the mean error rate on the left as a function of the number of samples for
   two stochastic training schedules. Standard error of the mean is shown as
   the shaded interval on the left and magnified in the right plot. It is
   evident that relatively few samples are needed for a reliable estimate of
   the mean error. The mean and standard error was computed using 30
   repetitions for each sample count.}
\label{fig:sample_estimates}
\end{figure}

\paragraph{Stochastic inference needs few samples for a good estimate:}
Figure~\ref{fig:sample_estimates} shows the estimated accuracies as a function
of the number of forward passes per image. It is evident that relatively few
samples are enough for a good estimate of the mean. Compare Figure-11
of~\cite{dropout}, which implies $\sim{50}$ samples are required.

\paragraph{Increase in width leads to considerable performance improvements:}
The number of filters in a convolutional layer is its width.
Table~\ref{tab:cifar10width} shows that the performance of a 20 layer model
improves considerably as the width is increased both for the baseline ResNet v2
architecture as well as the models trained with Swapout. Swapout is better able
to use the available capacity than the corresponding ResNet with similar
architecture and number of parameters. Table~\ref{tab:cifar10_soa} compares
models trained with Swapout with other approaches on CIFAR-10 while
Table~\ref{tab:cifar100_soa} compares on CIFAR-100. On both datasets our
shallower but wider model compares well with 1001 layer ResNet model.

\paragraph{Swapout uses parameters efficiently:}
Persistently over tables~\ref{tab:cifar1020layerbase},~\ref{tab:cifar10width},
and~\ref{tab:cifar10_soa}, Swapout models with fewer parameters outperform
other comparable models. For example, Swapout v2(32) $W \times 4$ gets 4.76\%
error with 7.43M parameters in comparison to the ResNet version at 4.91\% with
10.2M parameters.

\paragraph{Experiments on CIFAR-100 confirm our results:}
Table~\ref{tab:cifar100_soa} shows that Swapout is very effective as it
improves the performance of a 20 layer model (ResNet Ours) by more than 2\%.
Widening the network and reducing the stochasticity leads to further
improvements. Further, a wider but relatively shallow model trained with
Swapout (22.72\%; 7.46M params) is competitive with the best performing, very
deep (1001 layer) latest ResNet model (22.71\%;10.2M params).

\begin{table}[t]
  \caption{Wider swapout models work better. We evaluate the effect of
  increasing the number of filters on CIFAR-10.
  ResNets~\cite{resnets} contain three groups of layers with all convolutional
  layers in a group containing equal number of filters. We indicate the number
  of filters in each group as a tuple below and report the performance with
  deterministic as well as stochastic inference with 30 samples. For each size,
  model trained with Swapout outperforms the corresponding ResNet model.}
  \label{tab:cifar10width}
  \centering
  \begin{tabular}{llllll}
    \toprule
    Model & Width & \#Params & ResNet v2 & \multicolumn{2}{c}{Swapout} \\
    \cmidrule{4-5}
     & & & & Deterministic & Stochastic \\
    \midrule
    Swapout v2 (20) $W \times 1$  & (16, 32,  64) & 0.27M & 8.27 & 8.58 & 7.92 \\
    Swapout v2 (20) $W \times 2$  & (32, 64,  128)& 1.09M & 6.54  & 6.40 & 5.68 \\
    Swapout v2 (20) $W \times 4$  & (64, 128, 256)& 4.33M & 5.62 & 5.43 & 5.09 \\
    \midrule
    Swapout v2 (32) $W \times 4$  & (64, 128, 256)& 7.43M & \textbf{5.23} & \textbf{4.97} & \textbf{4.76} \\
    \bottomrule
  \end{tabular}
\end{table}

\begin{table}[t]
  \caption{Swapout outperforms comparable methods on CIFAR-10. Note that a 32
  layer wider model performs competitively in comparison to a 1001 layer ResNet
  model.}
  \label{tab:cifar10_soa}
  \centering
  \begin{tabular}{lll}
    \toprule
    Method   & \#Params & Error(\%)   \\
    \midrule
    DropConnect~\cite{dropconnect} & - & 9.32 \\
    NIN~\cite{netinnet}            & - & 8.81 \\
    FitNet(19)~\cite{fitnets}      & - & 8.39  \\
    DSN~\cite{lee2014deeply}       & - & 7.97        \\
    Highway\cite{highwaynet}       & - & 7.60 \\
    \midrule
    ResNet v1(110)~\cite{resnets}  & 1.7M & 6.41 \\
    Stochastic Depth v1(1202)~\cite{stochdepth} & 19.4M & 4.91 \\
    SwapOut v1(20) $W \times 2$    & 1.09M & 6.58 \\
    \midrule
    ResNet v2 (1001)~\cite{resnetsv2} & 10.2M & 4.92 \\
    SwapOut v2(32) $W \times 4$    & 7.43M & \textbf{4.76} \\
    \bottomrule
  \end{tabular}
\end{table}

\begin{table}[t]
  \caption{Swapout is strongly competitive with the best methods on CIFAR-100, and uses parameters efficiently in comparison. A 20 layer model (Swapout
  v2 (20)) trained with Swapout improves upon the corresponding 20 layer ResNet
  model (ResNet v2 Ours (20)). Further, a 32 layer wider but much shallower
  model performs competitively in comparison to a 1001 layer ResNet model (last
  row).}
  \label{tab:cifar100_soa}
  \centering
  \begin{tabular}{lll}
    \toprule
    Method     & Error(\%)   \\
    \midrule
    NIN~\cite{netinnet}                             & - & 35.68 \\
    DSN~\cite{lee2014deeply}                        & - & 34.57 \\
    FitNet~\cite{fitnets}                           & - & 35.04 \\
    Highway~\cite{highwaynet}                       & - & 32.39 \\
    \midrule  
    ResNet v1 (110)~\cite{resnets}                  & 1.7M  & 27.22 \\
    Stochastic Depth v1 (110)~\cite{stochdepth} & 1.7M  & 24.58 \\
    ResNet v2 (164)~\cite{resnetsv2}                & 1.7M  & 24.33 \\
    ResNet v2 (1001)~\cite{resnetsv2}               & 10.2M & \textbf{22.71} \\
    \midrule
    ResNet v2 Ours (20)            $W \times 2$     & 1.09M  & 28.08 \\
    \midrule
    SwapOut v2 (20)(Linear(1,0.5)) $W \times 2$     & 1.10M  & 25.86  \\
    SwapOut v2 (56)(Linear(1,0.5)) $W \times 2$     & 3.43M  & 24.86  \\
    SwapOut v2 (56)(Linear(1,0.8)) $W \times 2$     & 3.43M  & 23.46  \\
    SwapOut v2 (32)(Linear(1,0.8)) $W \times 4$     & 7.46M  & \textbf{22.72}  \\
    \bottomrule
  \end{tabular}
\end{table}

\vspace{-0.5mm}



\section{Discussion and future work}

Swapout is a stochastic training method that shows reliable improvements in
performance and leads to networks that use parameters efficiently. Relatively
shallow swapout networks give comparable performance to extremely deep residual
networks.

We have shown that different stochastic training schedules produce different
behaviors, but have not searched for the best schedule in any systematic way.
It may be possible to obtain improvements by doing so. We have described an
extremely general swapout mechanism.  It is straightforward using
equation~\ref{eq:swapoutgeneral} to apply swapout to inception
networks~\cite{googlenet} (by using several different functions of the input
and a sufficiently general form of convolution); to recurrent convolutional
networks~\cite{recurrentcnn} (by choosing $F_i$ to have the form $F \circ F
\circ F \ldots$); and to gated networks.  All our experiments focus on
comparisons to residual networks because these are the current top performers
on CIFAR-10 and CIFAR-100. It would be interesting to experiment with other
versions of the method.

As with dropout and batch normalization, it is difficult to give a crisp
explanation of why swapout works. We believe that our results support the idea
that swapout causes some form of improvement in the optimization process. This
is because relatively shallow networks with swapout reliably work as well as or
better than quite deep alternatives; and because swapout is notably and
reliably more efficient in its use of parameters than comparable deeper
networks. Unlike dropout, swapout will often propagate gradients while still
forcing units not to co-adapt. Furthermore, our swapout networks involve some
form of tying between layers.  When a unit sometimes sees layer $i$ and
sometimes layer $i-j$, the gradient signal will be exploited to encourage the
two layers to behave similarly. The reason swapout is successful likely
involves both of these points.

\paragraph{Acknowledgments:}
This work is supported in part by ONR MURI Awards N00014-10-1-0934 and
N00014-16-1-2007. We would like to thank NVIDIA for donating some of the GPUs
used in this work.

{\small
\bibliographystyle{ieee}
\bibliography{bib_all}
}

\end{document}